\documentclass{article}

\usepackage[numbers]{natbib}
\usepackage{amsmath}
\usepackage{graphicx}
\usepackage{caption}

\usepackage[preprint]{neurips_2021}

\usepackage[utf8]{inputenc} 
\usepackage[T1]{fontenc}    
\usepackage{hyperref}       
\usepackage{url}            
\usepackage{booktabs}       
\usepackage{amsfonts}       
\usepackage{nicefrac}       
\usepackage{microtype}      
\usepackage{subcaption}
\usepackage{multirow}
\title{Generative Models for 3D Point Clouds}

\author{%
    Lingjie Kong \thanks{All authors contributed equally} \\
    Department of Computer Science \\
    Stanford University\\
    \texttt{ljkong@stanford.edu} 
    \And
    Pankaj Rajak  \\
    Department of Computer Science \\
    Stanford University\\
    \texttt{prajak7@stanford.edu} 
    \And
    Siamak Shakeri \\
    Department of Computer Science \\
    Stanford University\\
    \texttt{siamaks@stanford.edu} 
}
\begin{document}

\maketitle

\begin{abstract}
Point clouds are rich geometric data structures, where  their three dimensional structure offers an excellent domain for understanding the representation learning and generative modeling in 3D space. In this work, we aim to improve the performance of point cloud latent-space generative models by experimenting with transformer encoders, latent-space flow models, and autoregressive decoders. We analyze and compare both generation and reconstruction performance of these models on various object types.
\end{abstract}

\section{Introduction}
Point clouds are rich geometric data structures, where  their three dimensional structure offers an excellent domain for understanding the representation learning and generative modeling in 3D space.  Learning generative models for point clouds is an emerging field, where the models are trained to be able to generate a set of points in 3D space that resemble the object which is has been previously trained on. Point clouds as a representation are permutation invariant. Any reordering of the rows of the point cloud matrix should yield a point cloud that represents the same shape. This is currently achieved using \textit{maxpooling}, which is a \textit{symmetrictly} permutation invariant function  during encoding. However, this ignores point relationships which are achieved similarly through Convolution Neural Networks in image application. Transformer encoders \cite{yuan2018iterative}, \cite{yang2019modeling} can be used to encode the relationship between points of a point cloud while preserving permutation invariant property. We hypothesize that using such encoder enables capturing more complex point relationships, thus improving reconstruction and training performance of the models. 
Another issue with current point cloud models is the assumption of approximating the posterior by mixture of Gaussian. This could hinder the generative performance of the model if the true latent-space distribution of the model is non-Gaussian. We propose using normalizing flow models \cite{normalizingflow} in the latent space by transforming simple Gaussian latent space representations into more complex ones \cite{yang2019pointflow}.

Another aspect of point clouds that we explore is improving the decoding process from the latent space code. A multi-layer perceptron decoder, such as the one used in \cite{achlioptas2017learning}, might not be powerful enough, as it generates all the point clouds jointly at once. We propose using a progressive decoder model \cite{heljakka2018pioneer}. Through decoding 3D point cloud in a progressive manner, we can generate the points in the point cloud by conditioning on the already generated points. We hypothesize that this would facilitate generating more complex objects.

The rest of this paper is organized as follows: We discuss the related work in the next section, which is followed by problem statement and technical approach. In section 4, the dataset used and the experimental results are mentioned. This section is followed by analysis of the results and future work.

\section{Related Works}
The initial interest on 3D point cloud was on classification, segmentation, and object detection \cite{qi2017pointnet}, \cite{qi2017pointnet++}, \cite{zhou2018voxelnet}. In order to make sure the input size is consistent, the input is either chosen from a fixed number of points sampling from a point cloud or Voxel representation. Authors in \cite{oliva2018transformation} have proposed transformation autoencoder networks, where they keep point clouds permutation invariant through maxpooling. Studies are performed on designing more complicated permutation-invariant functions such as deep set \cite{zaheer2017deep} with specific neural network architectures. Others have explored spherical harmonic functions \cite{thomas2018tensor}. Besides designing more powerful permutation-invariant functions, there has been work done on developing a hybrid GAN-VAE model to better represent the latent space \cite{achlioptas2017learning}. Flow models have been explored to solve 3D point generation in \cite{yang2019pointflow}. 

The current state of art method [1] to build generative model for point cloud is based on a hybrid approach consisting of Variational autoencoder (VAE) and Generative adversarial network (GAN), where the latent space of VAE is used as an input to the GAN model. 

As previously mentioned, encoders used in point cloud deep models should be permutation invariant. Previous works on generative models of point clouds consist of a variant of multi-layer perceptron (MLP) [1], which is shared among all points present inside the point set. Recently, a new method has been proposed to design filters for 3D point clouds, which is not only translation and permutation invariant, but also equivalent to rotation [3]. These filters are based on the product of radial basis functions and spherical harmonics. The parameters of spherical harmonics in these filters captures local neighborhood information of a point in hierarchical manner. For example, these filters can encode radial and angular local neighborhood of each points separately.


\section{Problem Statement and Technical Approach}

    \subsection{Definition}
    A point cloud is a set of points $S$, where each $s \in R^3$ is a tuple that determines the x,z, and z coordinates of the surface of a 3D object in the Euclidean space. We assume $|S|=N$.
    
    \textbf{Density Estimation:} Given a collection of point clouds $\{S_i\}$ of a certain object, such as airplane, learn a parameterized density function $p_{\theta}(S)$.
    
    \textbf{Generation:} Create $N$ tuples of point that jointly represent the surface of a 3D object. \\
    \textbf{Reconstruction:} Given a set of $N$ points representing the surface of an object, conditionally generate $N$ tuples that resemble the same object.
    
\begin{figure}[ht]
        \begin{center}
            \includegraphics[width=.50\textwidth]{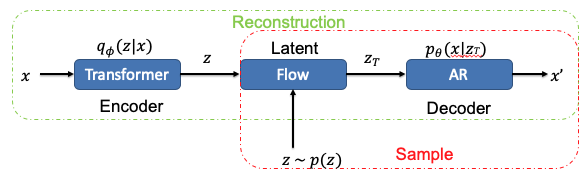}
        \end{center}
        \caption{VAE Architecture}
        \label{VAE Architecture}
    \end{figure}

    Figure~\ref{VAE Architecture} depicts all of our proposed improvements to a baseline VAE model when using transformer as encoder, flow model for latent space, and autoregressive decoder in which :
    \begin{itemize}
        \item $x$ is the 3D point clouds input data. 
        \item $q_{\phi}(z|x)$ is the encoder approximated posterior distribution. 
        \item $p(z)$ is  the prior distribution. 
        \item $z$ is the original latent representation from encoder as mixture of Gaussian.
        \item $z_T$ is the transformed latent representation after applying flow model. 
        \item $p_{\theta}(x|z_T)$ is the decoder likelihood distribution. 
        \item $x'$ is the output reconstruction/sample 3d point clouds.
    \end{itemize}

    \subsection{VAE: Variational Autoencoder Model}
\begin{figure}[ht]
        \begin{center}
            \includegraphics[width=.50\textwidth]{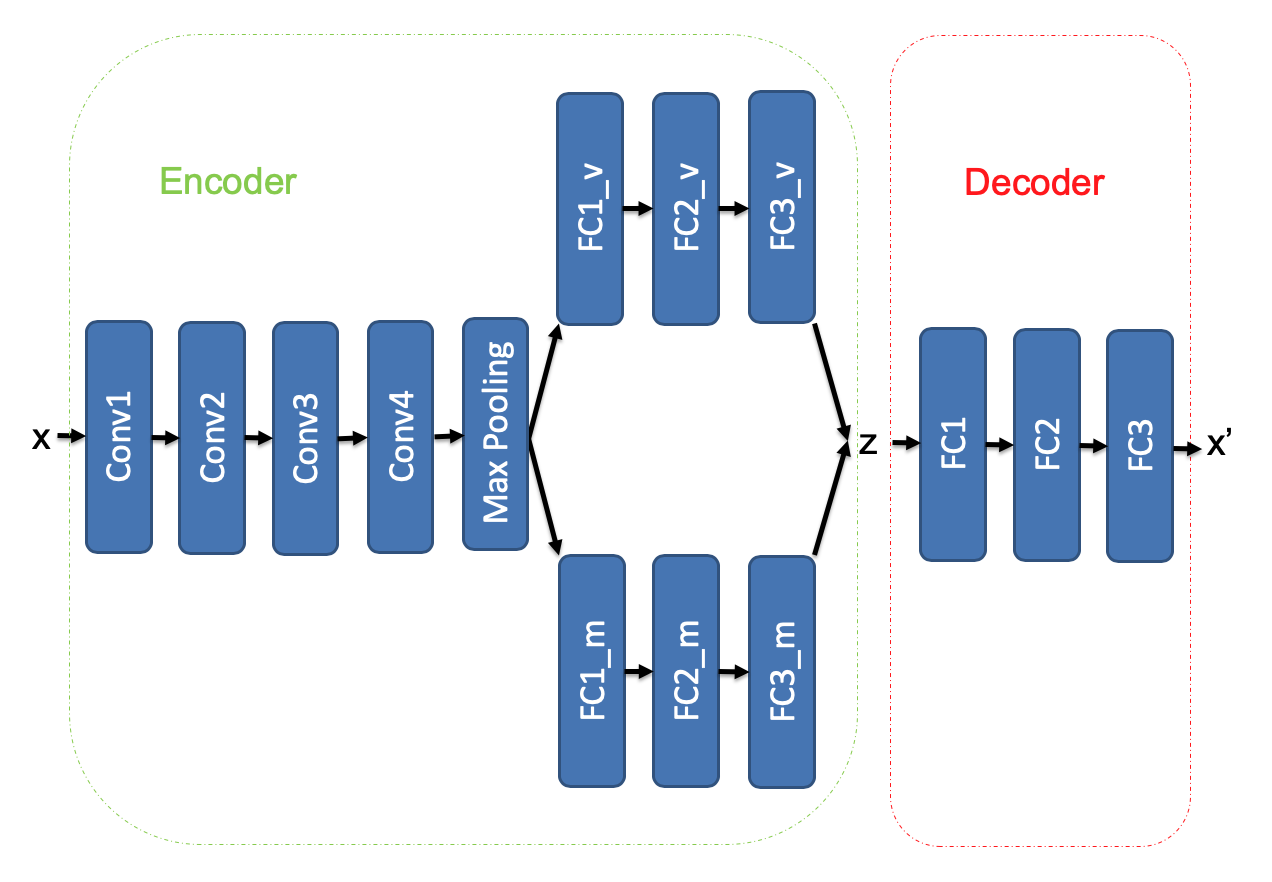}
        \end{center}
        \caption{Baseline VAE Network}
        \label{Baseline VAE Network}
    \end{figure}

    Baseline VAE which contains 4 conv layers, 1 max pooling, and 3 fully connected layers for the encoder and 3 fully connected layers for the decoder is shown in Figure~\ref{Baseline VAE Network}.

    The prior is modeled as uniform Gaussian $p(\textbf{z})$
    $$p(\textbf{z}) = \mathcal{N}(\textbf{x}|0, I)$$
    
    Lower bound can be maximized to the marginal log-likelihood to obtain an expression known as the \textbf{evidence lower bound} (ELBO) with encoder network $q_\phi(\textbf{z} | \textbf{x})$  and decoder network $p_\theta(\textbf{x}|\textbf{z})$. 
    $$\log p_\theta(\textbf{x}) \geq \mathrm{ELBO}(\textbf{x}; \theta, \phi) = \mathcal{L}(\textbf{x}; \theta, \phi)$$
    $$ = \mathbb{E}_{q_\phi(\textbf{z} | \textbf{x})}[\log p_\theta(\textbf{z},\textbf{x}) - \log(q_{\phi}(\textbf{z}|\textbf{x}))]$$
    $$ = \mathbb{E}_{q_\phi(\textbf{z} | \textbf{x})}[\log p_\theta(\textbf{x}|\textbf{z})] - D_{\text{KL}}(q_\phi(\textbf{z} | \textbf{x})||p(\textbf{z})))$$
    
    
    During training, the original 3D points are feed as input for the encoder network to obtain mean ($\mu_\phi(\textbf{x})$) and covaraince ($\mathrm{diag}(\sigma^2_\phi(\textbf{x})$). Then $z$ is sample as below.
    $$q_\phi(\textbf{z}|\textbf{x}) = \mathcal{N}(\textbf{z}|\mu_\phi(\textbf{x}),  \mathrm{diag}(\sigma^2_\phi(\textbf{x})))$$
    
    Assuming $p_\theta(\textbf{x}|\textbf{z})$ is a Gaussian distribution with output as mean and constant covariance, then the reconstruction loss can be written as the chamber's distance loss between the mean of sample $x'\sim p_\theta(\textbf{x}|\textbf{z})$ and ground truth x.
    $$d_{CD}(S_1,S_2) = \sum_{x\in S_1}min_{x'\in S_2}||x-x'||_2 + \sum_{x'\in S_2}min_{x\in S_1}||x-x'||_2$$
    
    
    During Sampling, one can sample from $p(\textbf{z})$ and feed the sample as input for the decoder network $p_\theta(\textbf{x}|\textbf{z})$ to generate 3D points.

    \subsection{Transformer Encoder}
    The baseline VAE uses 1D convolution layers to encode the input points. Therefore, encoding of a point does not depend on the rest of the points in the network. This could potentially degrade the performance of the models by not providing rich encodings.
    
    We propose using self-attention encoder on top of the convolution layers to augment the encoder of the baseline VAE. We use the transformer encoder of \cite{vaswani2017attention}, which is very popular in natural language processing. 
    
\begin{figure}
\centering
\begin{minipage}{.5\textwidth}
  \centering
  \includegraphics[width=.5\linewidth]{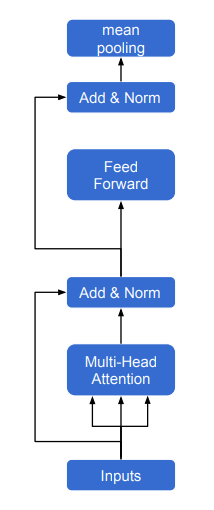}
  \caption*{Transformer Encoder}
  \label{fig:sub1}
\end{minipage}%
\begin{minipage}{.5\textwidth}
  \centering
  \includegraphics[width=1\linewidth]{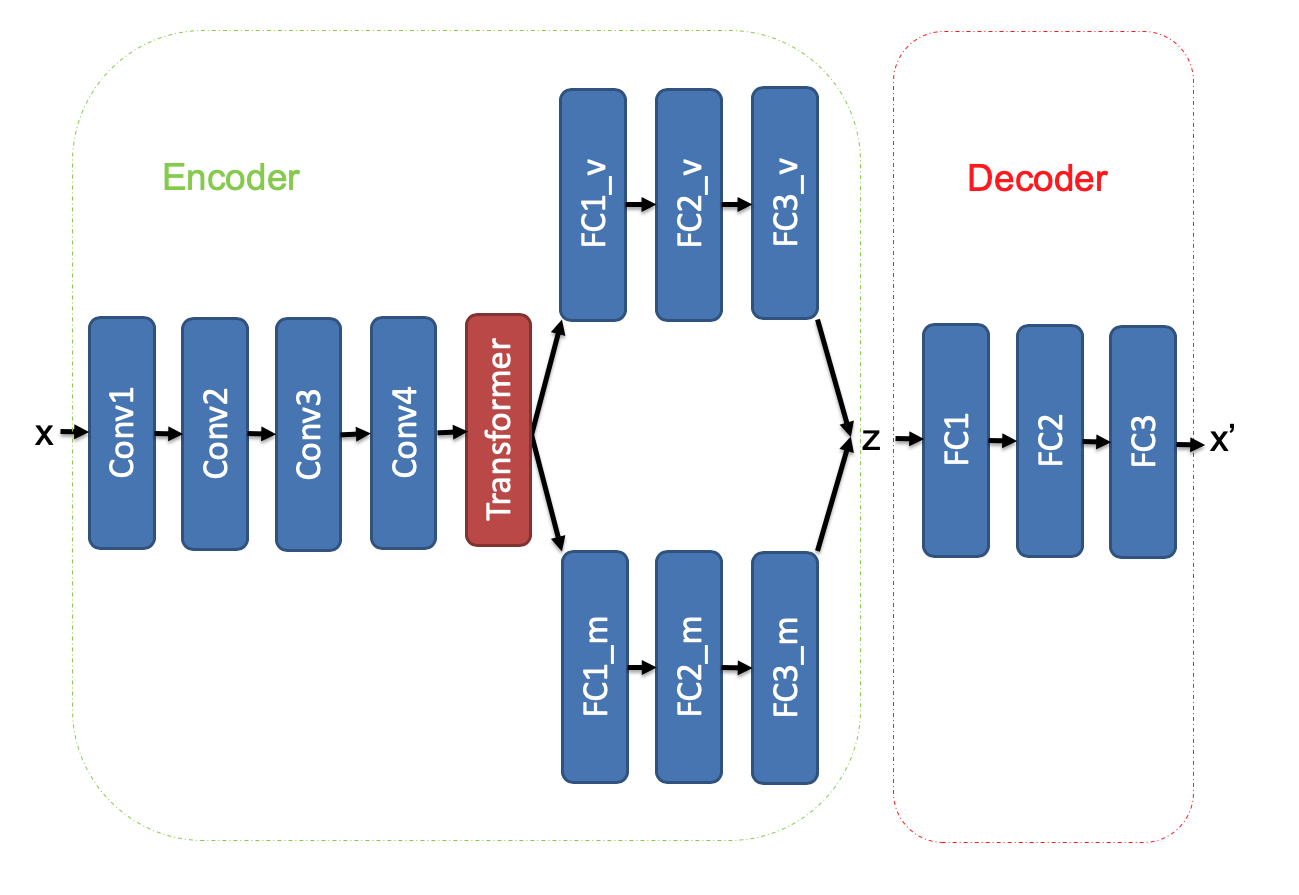}
  \caption*{VAE network with Transformer Encoder}
  \label{fig:sub2}
\end{minipage}
\caption{Transformer Encoder design for Point Clouds}
\label{fig:transformer}
\end{figure}

    Figure \ref{fig:transformer} shows the transformer encoder, and its application to the VAE model.
    
Different from \cite{vaswani2017attention}, we do not employ positional encoding in the transformer encoder. This is due to the fact that absolute position of points fed to the model does not imply their distance.

    \subsection{VAE with Flow Latent Space using IAF}
\begin{figure}[ht]
        \begin{center}
            \includegraphics[width=.50\textwidth]{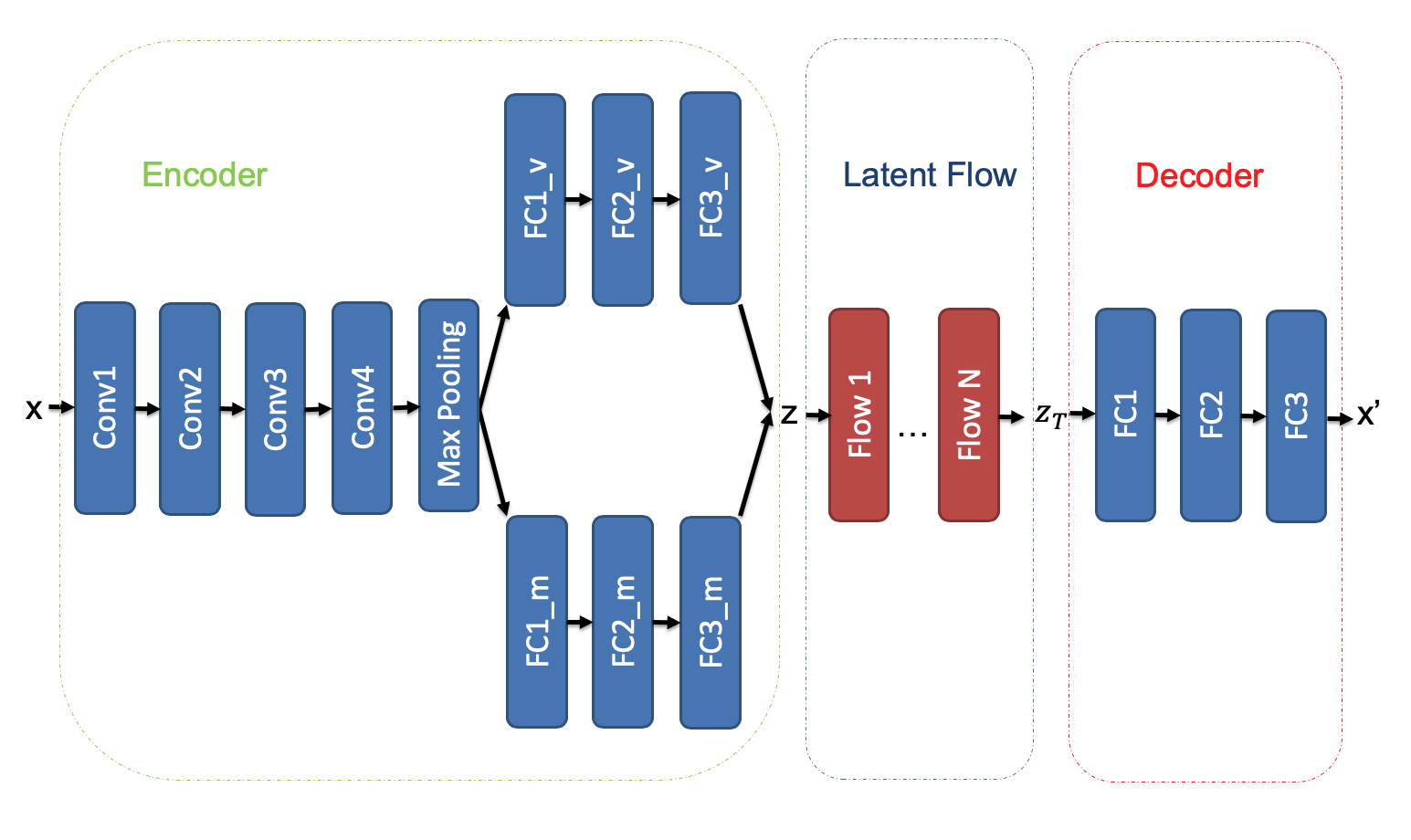}
        \end{center}
        \caption{Flow Latent Space VAE Network}
        \label{Flow Latent Space VAE Network}
    \end{figure}

    An inverse autoregressive flow model (IAF) \cite{kingma2016improved} is used to model the posterior $q(z|X)$, where the output of the encoder is feed into an invertible IAF layer. The output of the IAF layer is given as an input to the decoder. Its structure can be seen in Figure~\ref{Flow Latent Space VAE Network}.
    
    The VAE loss function using flow in the latent space that is optimized as follows.
    Recall the ELBO is written as below for baseline VAE.
    $$ \mathcal{L}(\textbf{x}; \theta, \phi) = \mathbb{E}_{q_\phi(\textbf{z} | \textbf{x})}[\log p_\theta(\textbf{z},\textbf{x}) - \log(q_{\phi}(\textbf{z}|\textbf{x}))] $$
    For latent flow VAE, the loss function will be.
    $$ \mathcal{L}(\textbf{x}; \theta, \phi) = \mathbb{E}_{q_\phi(\textbf{z}_T | \textbf{x})}[\log p_\theta(\textbf{z}_T,\textbf{x}) - \log(q_{\phi}(\textbf{z}_T|\textbf{x}))] $$
    $$ = \mathbb{E}_{q_\phi(\textbf{z}_T | \textbf{x})}[\log p_\theta(\textbf{x}|\textbf{z}_T) + \log( p_\theta(\textbf{z}_T)) - \log(q_{\phi}(\textbf{z}_T|\textbf{x}))] $$
    in which the prior $p_\theta(\textbf{z}_T))$ is modeled by a Gaussian distribution and log posterior $\log(q_{\phi}(\textbf{z}_T|\textbf{x}))$ according to \cite{kingma2016improved} is as below
    $$ \log(q_{\phi}(\textbf{z}_T|\textbf{x})) = -\sum_{i=1}^{D}(\frac{1}{2}\epsilon_i^2+\frac{1}{2}\log(2\pi) + \sum_{t=0}^T(log(\sigma_t,i))) $$
    
    \subsection{VAE with Autoregressive Decoder using NADE}
\begin{figure}[ht]
        \begin{center}
            \includegraphics[width=.50\textwidth]{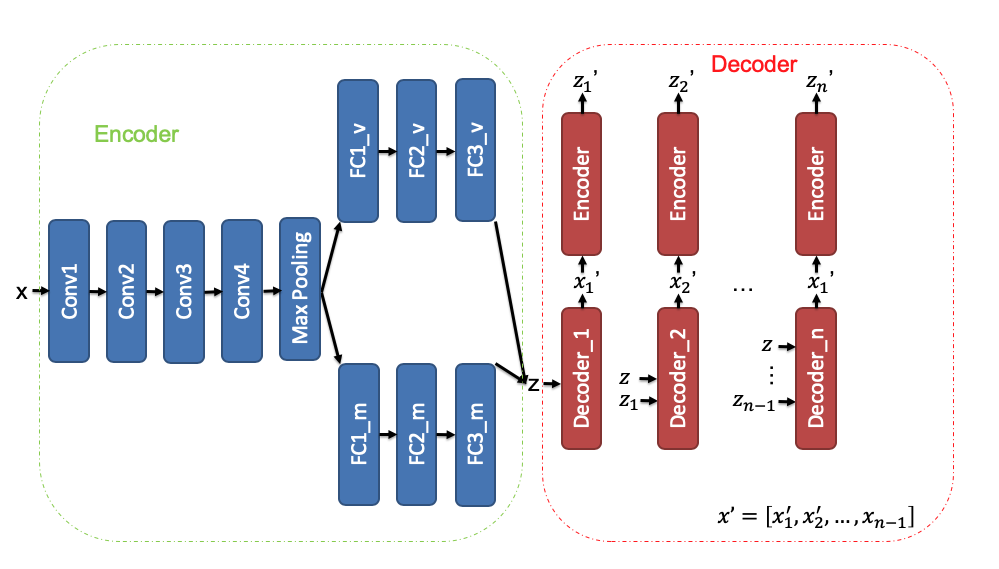}
        \end{center}
        \caption{Autoregressive decoder VAE Network}
        \label{Autoregressive decoder VAE Network}
    \end{figure}

    The decoder is modeled in an autoregressive manner following a Neural Autoregressive Distribution Estimation (NADE) \cite{uria2016neural} structure. Its structure can be shown in Figure~\ref{Autoregressive decoder VAE Network}.
    
    The conditional probability of our decoder can be written as
    $$P(X|z) = P_1(x_1|z)P_2(x_2|z,z_{x_1})P_3(x_3|z,z_{x_1},z_{x_2})...P_n(x_n|z, z_{x_1}, z_{x_2}, ..., z_{x_{n-1}}) $$
    
    A point cloud $X$ is partitioned into n groups, in which each conditional probability, $P_n(x_i|z, z_{x_1}, z_{x_2}, ..., z_{x_{i-1}})$ is modeled by a multi-layer perceptron,which joint decode $x_i$ given $z, z_{x_1}, z_{x_2}, ..., z_{x_{i-1}}$. Here, $z$ is sampled from posterior encoder distribution $q(z|X)$. $z_{x_i}$ is m-dimension latent representation of points from group $x_i$ using PointNet architecture,which is shared amoung all groups. For our model, $n=8$ and a total number of $2048$ points is divided into $8$ groups of $256$ points. Latent space dimension $m$ is 128. Furthermore, the auto regressive decoding of points requires an ordering in which the points needs to generated. Hence, during decoding points are sorted according to one of the x,y,z axis and then partitioned into n groups, where each group is decoded by one NADE block.

    \subsection{Point Cloud generation using Masked Autoregressive Density Estimation (MADE)}
    Autoregressive models (AR) offer a flexible way to estimate the density of a data distribution, which can later be used not only to compute log likelihood of new data points but also generate samples from it. However, these models require an ordering of the input data. To build an AR model for point cloud objects, we have first sorted the objects along one of its axes and then used the same axis during the generation. Here, each point is modeled by a Gaussian distribution, where its parameters - $\mu_i$ and $\sigma_i$- are modeled using 5 MADE layers. The log-likelihood of each object is given as 
    $$logP(X)=P(x_1)P(x_2|x_1)P(x_3|x_2,x_1)...P(x_n|x_1,x_2,...x_{n-1})$$
    where, $$P(x_i|x_1,x_2,...x_{i-1})=\frac{1}{(2\pi \sigma_i^2 )^{\frac{1}{2}}} exp^{(-(x_i-\mu_i)^2/(2\sigma_i^2))}$$
    
    During training, we have also constrained the values of $\sigma_i$ by adding an L1 regularization such that $\sum{\sigma_i}$ is minimized as it has shown to increase the sample quality during generation. Figure 9 shows the samples generated by our MADE model after training for airplane. It can be observed that the quality of the samples generated are not as good as baseline VAE, but they still captures some of the basic features of airplanes such as wings of the airplane. 
    
    As a future research direction our goal is to combine VAE with MADE decoder. Here, the auto-regressive property of MADE will allow tractable computation of P(X|z) using mixture of Gaussian and the latent variable z from VAE will provide global property of the objects to the decoder. 

\section{Experiment and Result}
    \subsection{Dateset}
    ShapeNet (\cite{shapenet}) includes a repository of 3D shapes. ShapeNet Core which covers 55 common object categories with about 51,300 unique 3D models are used as the input. Point clouds are created by uniformly sampling from these shapes to convert them to point clouds, as suggested in \cite{achlioptas2017learning}. This dataset is used for experimentations.

    Besides ShapeNet, QM9 dataset (\cite{Ramakrishnan2014QuantumCS}) which is comprised of 134,000 molecules and MD17 molecular dynamics forces (\cite{Chmielae1603015}) are other 3D datasets that can be employed in our work.
    
    Another dataset for point clouds is KITTI benchmark suite, which includes 3D Velodyne point clouds.

    \subsection{Evaluation}
        One of the challenges when evaluating point clouds is that the metrics need to be permutation invariant. The following distance metrics are proposed in literature for point clouds:
        \begin{itemize}
            \item Earth Mover's Distance(EMD):
            \begin{equation}
                d_{EMD}(S_1,S_2) = \min_{\phi:S_1\rightarrow S_2}\sum_{x\in S_1}\lVert x-\phi(x)\rVert_2
            \end{equation}
            \item Chamfer's Distance:
            \begin{equation}
                d_{CD}(S_1,S_2) = \sum_{x\in S_1}min_{y\in S_2}||x-y||_2 + \sum_{y\in S_2}min_{x\in S_1}||x-y||_2
            \end{equation}
        \end{itemize}
        
        Both distance metrics are differentiable. Chamfer's Distance is more computationally efficient than EMD. We have used CD in our work.
        
        \cite{achlioptas2017learning} proposes a number of evaluation metrics to measure the similarity of two points clouds A and B using the distance metrics mentioned above. They are as follows:
        \begin{itemize}
            \item Jensen-Shannon Divergence: This metric measures how close the points clouds of one set are to the point clouds in another set in the Euclidean 3D space. The closer JSD is to 0, the less divergence there is.
            \item Coverage: It measures the fraction of points in point cloud B that are closest point to a point in A. This metrics intuitively measures how many points in B are represented in A. The closer coverage is to 1, the better coverage there is.
            \item Minimum Matching Distance(MMD): This is similar to Coverage, except relying on the average of the minimum distances of matched pairs. The smaller MMD is, the better matching there is.
        \end{itemize}
    
    \subsection{Result}
    \begin{itemize}
        \item \textbf{VAE:} Baseline VAE 
        \item \textbf{VAE+Trx:} \textbf{Tr}ansformer Encoder 
        \item \textbf{VAE+AR:} \textbf{A}utoregressive \textbf{D}ecoder
        \item \textbf{VAE+Flow:} Latent-Space \textbf{Flow}
    \end{itemize}
    Table \ref{chair table} Chair, Table \ref{car table} Car, and Table \ref{airplane table} airplane show reconstruction and sample results for VAE, VAE+Trx, VAE+Flow, and VAE+AR models separately during both train and test. Baseline VAE model exhibits competitive performance while there is not a specific augmented model consistently out-performing others among all category using JSD, coverage, and MMD as metrics.
    
\begin{figure}[ht]
        \begin{center}
            \includegraphics[width=.95\textwidth]{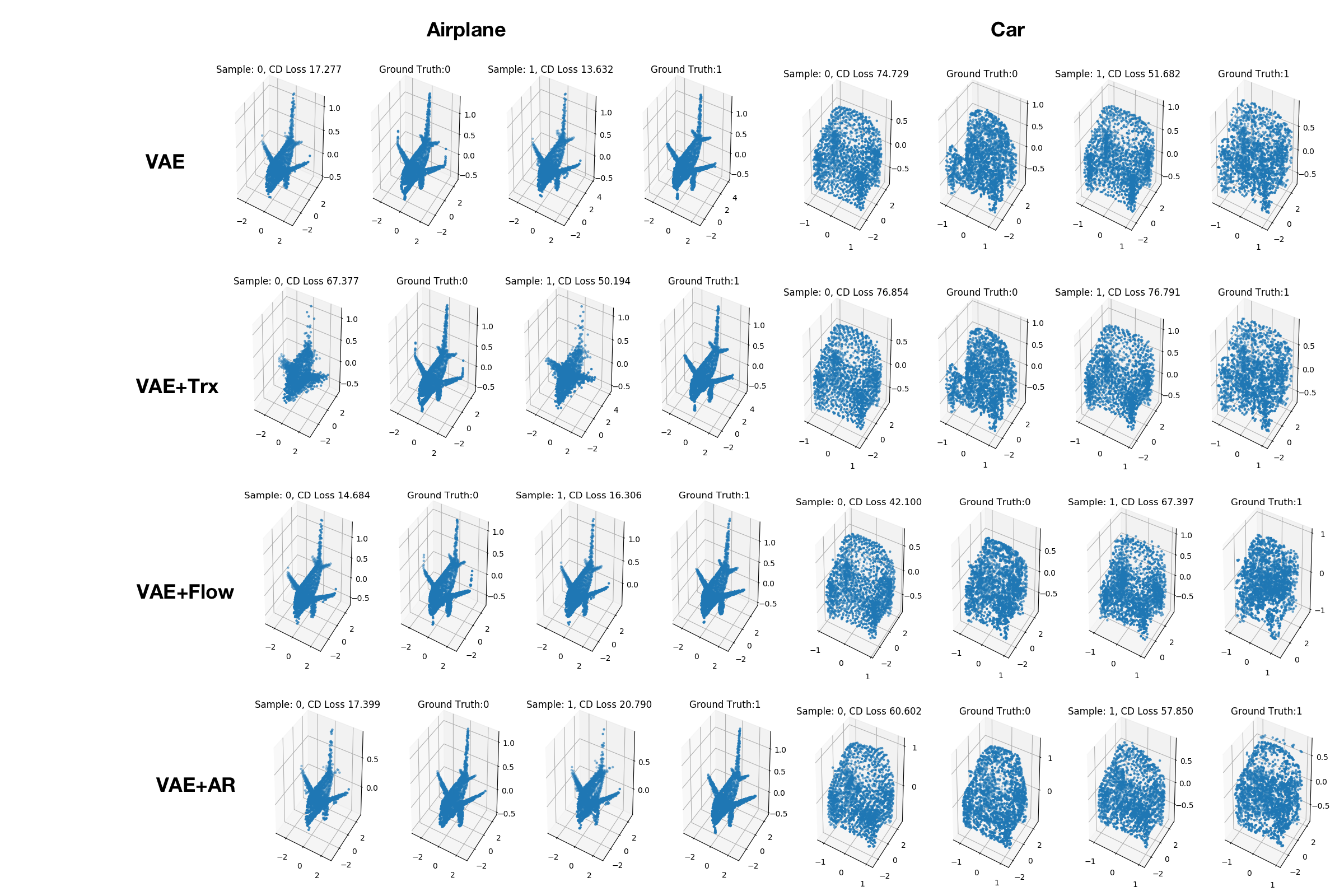}
        \end{center}
        \caption{Reconstruction Result}
        \label{Reconstruction Result}
    \end{figure}

    Figure \ref{Reconstruction Result} shows the reconstruction and ground truth result using airplane and car dataset during training. Baseline VAE model, VAE+Flow, and VAE+AR can reconstruct both the overall shape and details while VAE+Trx reconstructs the overall shape, but fail to reconstruct the airlplane in the airplane dataset. For each reconstructed object, the its CD distrance from the ground truth object is shown in the top
    
\begin{figure}[ht]
        \begin{center}
            \includegraphics[width=.85\textwidth]{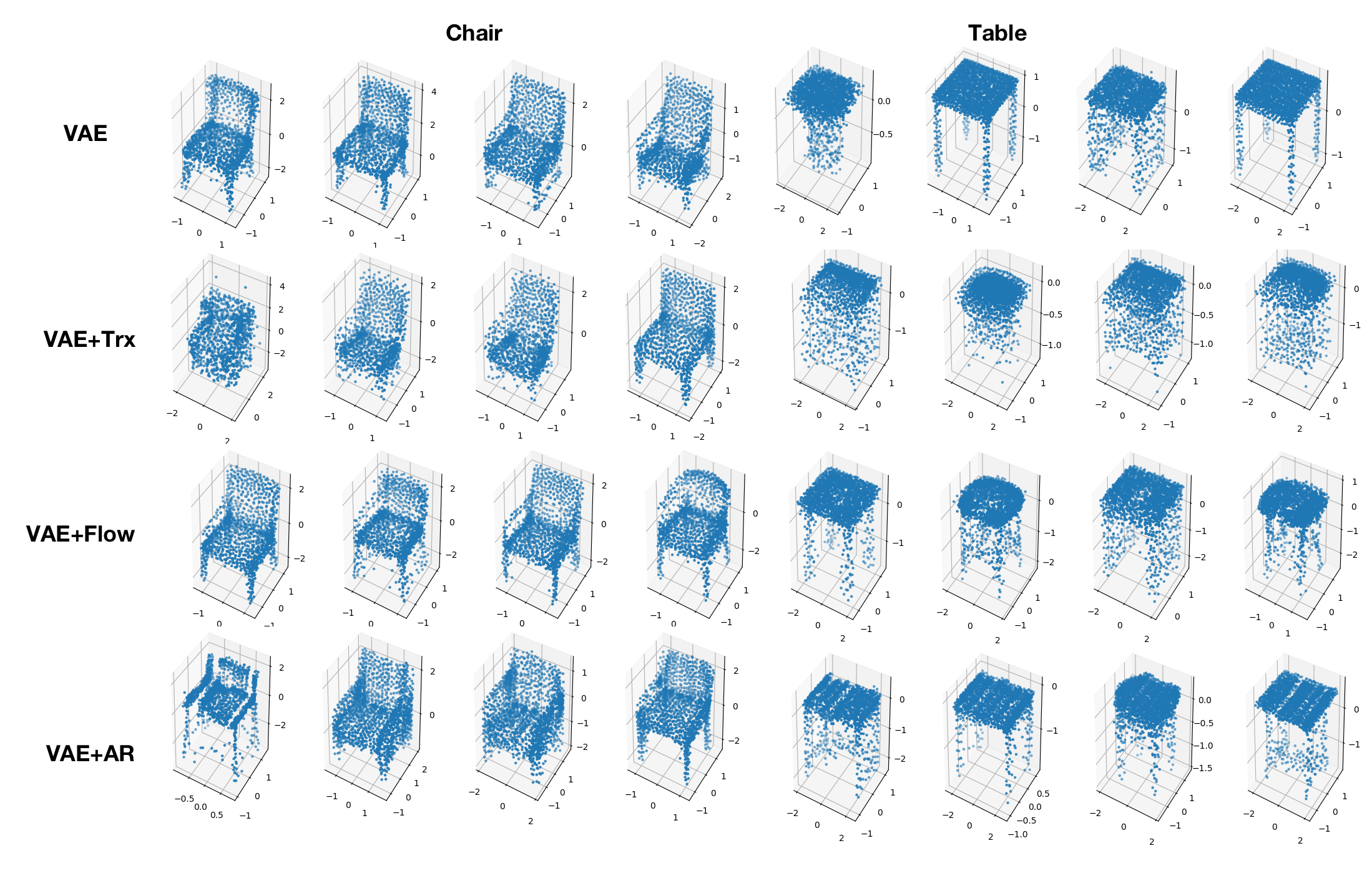}
        \end{center}
        \caption{Sample Result}
        \label{Sample Result}
    \end{figure}

    Figure \ref{Sample Result} shows the sample test results for chair and table. Baseline VAE model, VAE+Trx, VAE+Flow, VAE+AR can capture the overall shape for both chair and table. However, details such as chair and tables legs cannot be formed clearly during sample.

\begin{figure}[ht]
        \begin{center}
            \includegraphics[width=.85\textwidth]{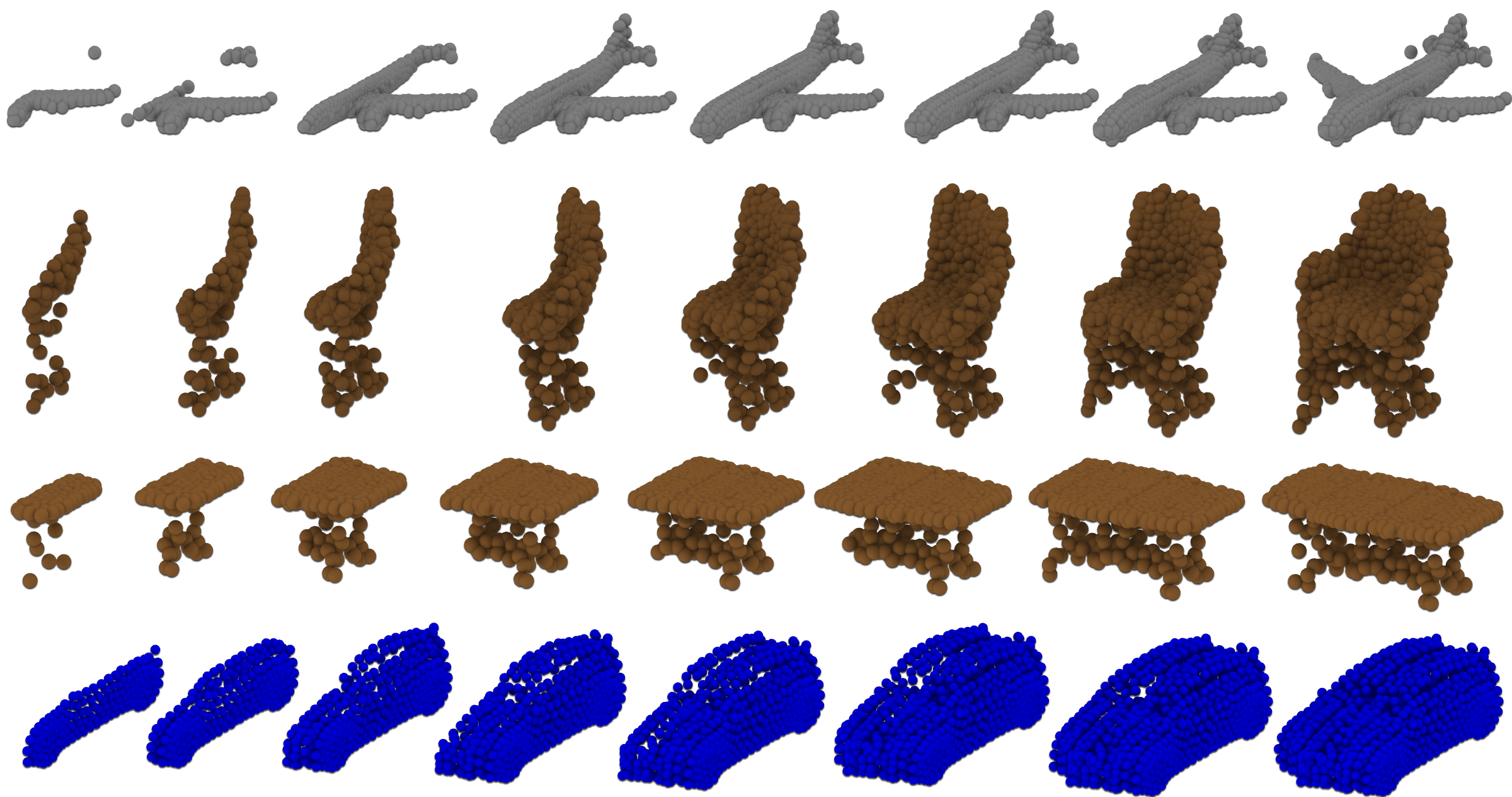}
        \end{center}
        \caption{Autoregressive Decoding Result}
        \label{Autoregressive Decoding Result}
    \end{figure}

    Figure \ref{Autoregressive Decoding Result} shows autoregresive decoding result to decode $2048$ points for a progressive manner with $256$ points per step for 8 steps in total.
    
\begin{figure}[ht]
        \begin{center}
            \includegraphics[width=.85\textwidth]{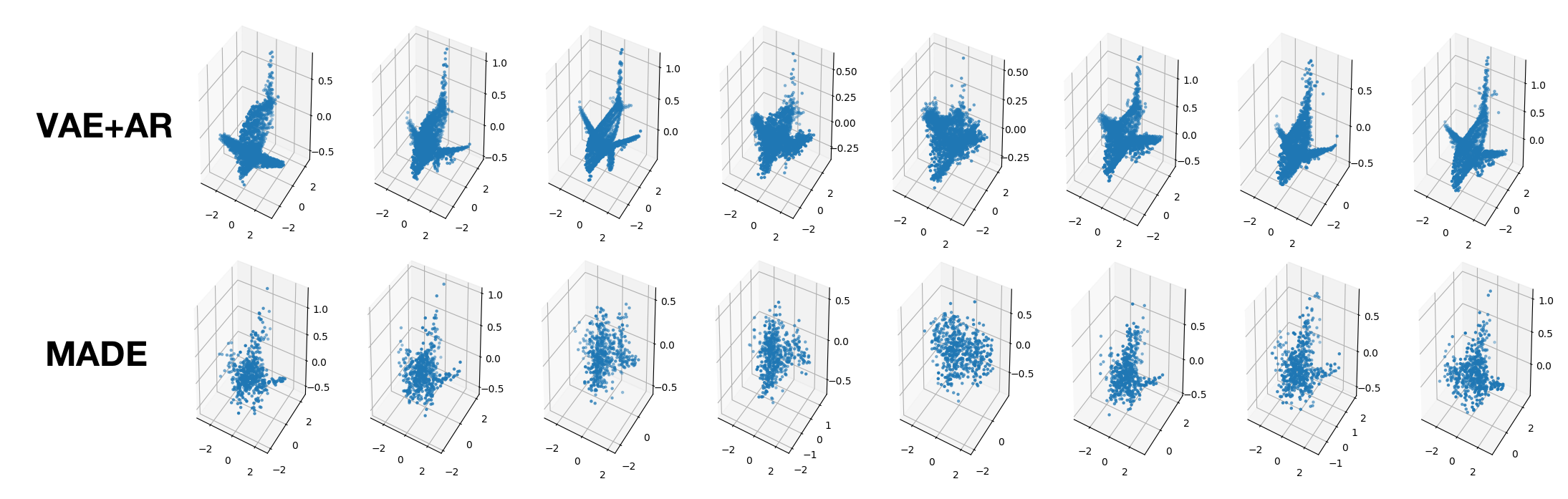}
        \end{center}
        \caption{VAE vs. MADE (VAE models are trained with 2048 points whereas MADE models are trained with 512 points)}
        \label{VAE vs. MADE}
    \end{figure}

    Figure \ref{VAE vs. MADE} shows the VAE model compared to the MADE. Unlike VAE models trained with 2048 points, MADE models are trained with 512 points to speed up the training and sampling of objects.

    \begin{table}
    \tiny
    \caption{\scriptsize{Reconstruction and Generation Evaluation Metrics of Chair}}
    \label{chair table}
    \begin{tabular}{ |c|c|c|c|c|c|c|c|c| } 
    \hline
    Samples&Model&Split&JSD$\downarrow$ &Coverage$\uparrow$ &MMD$\downarrow$&NELBO$\downarrow$&KL$\downarrow$&Reconst Loss$\downarrow$ \\ \hline
    \multirow{7}{*}{Reconstruction}&\multirow{3}{*}{Train}&VAE&0.08127&\textbf{0.85061}&\textbf{1.56782}&73.32367&17.90308&\textbf{55.42059}\\ 
    &&VAE+Trx&0.14206&0.49935&2.22028&102.69266&14.69619&87.99647 \\ 
    &&VAE+AR&\textbf{0.06992}&0.84866&1.7212&80.10754&18.6372&61.47034 \\ \cline{2-9}
    &\multirow{4}{*}{Test}&VAE&0.09089&0.79271&\textbf{2.32123}&101.79727&18.02342&\textbf{83.77386}\\  
    &&VAE+Trx&0.15491&0.51025&2.91711&126.22491&14.67373&111.55118 \\ 
    &&VAE+AR&\textbf{0.07546}&\textbf{0.79575}&2.43235&107.27012&18.69086&88.57926 \\ 
    &&VAE+Flow&0.10518&0.76765&2.40423&103.23891&17.17888&86.06003 \\ \hline
    
    \hline
    \multirow{7}{*}{Generation}&\multirow{3}{*}{Train}&VAE&0.10043&\textbf{0.45642}&3.0537&&& \\
    &&VAE+Trx&0.15628&0.3931&\textbf{2.84244}&&& \\
    &&VAE+AR&\textbf{0.09164}&0.42736&3.24288&&& \\  \cline{2-6}
    &\multirow{4}{*}{Test}&VAE&0.10925&\textbf{0.43508}&4.11022&&& \\
    &&VAE+Trx&0.16539&0.41989&\textbf{3.64259}&&& \\
    &&VAE+AR&\textbf{0.10296}&0.40015&4.2808&&& \\ 
    &&VAE+Flow&0.24696&0.20881&5.38606&&& \\ \hline
    \end{tabular}
    \end{table}
    \begin{table}
    \tiny
    \caption{\scriptsize{Reconstruction and Generation Evaluation Metrics of Car}}
    \label{car table}
    \begin{tabular}{ |c|c|c|c|c|c|c|c|c| } 
    \hline
    Samples&Model&Split&JSD$\downarrow$ &Coverage$\uparrow$ &MMD$\downarrow$&NELBO$\downarrow$&KL$\downarrow$&Reconst Loss$\downarrow$ \\ \hline
    \multirow{7}{*}{Reconstruction}&\multirow{3}{*}{Train}&VAE&0.05157&\textbf{0.40155}&\textbf{1.18744}&53.80332&8.42543&\textbf{45.37789} \\
    &&VAE+Trx&0.06903&0.31448&1.29159&58.98346&7.61572&51.36774 \\
    &&VAE+AR&\textbf{0.04167}&0.3991&1.23474&55.81075&8.33354&47.47721 \\ \cline{2-9}
    &\multirow{4}{*}{Test}&VAE&0.05594&0.46733&1.79432&78.51968&9.43197&69.08771 \\ 
    &&VAE+Trx&0.09031&0.33807&1.97529&91.12982&7.79197&83.33785 \\
    &&VAE+AR&\textbf{0.047}&\textbf{0.47301}&1.85738&79.25379&9.25665&69.99714 \\ 
    &&VAE+Flow&0.06515&0.46591&\textbf{1.76742}&77.22068&9.42193&\textbf{67.79875} \\
    \hline
    
    \hline
    \multirow{7}{*}{Generation}&\multirow{3}{*}{Train}&VAE&0.05342&\textbf{0.34418}&\textbf{1.36901}&&& \\
    &&VAE+Trx&0.07256&0.29251&1.39957&&& \\
    &&VAE+AR&\textbf{0.04651}&0.33157&1.44784&&& \\ \cline{2-6}
    &\multirow{4}{*}{Test}&VAE&0.07998&0.34801&2.28533&&& \\
    &&VAE+Trx&0.0972&0.31392&2.2712&&& \\
    &&VAE+AR&\textbf{0.06792}&\textbf{0.34943}&\textbf{2.24429}&&&\\ 
    &&VAE+Flow&0.14089&0.22443&2.79232&&& \\\hline
    \end{tabular}
    \end{table}
    \begin{table}
    \tiny
    \caption{\scriptsize{Reconstruction and Generation Evaluation Metrics of Airplane}}
    \label{airplane table}
    \begin{tabular}{ |c|c|c|c|c|c|c|c|c| } 
    \hline
    Samples&Model&Split&JSD$\downarrow$ &Coverage$\uparrow$ &MMD$\downarrow$&NELBO$\downarrow$&KL$\downarrow$&Reconst Loss$\downarrow$ \\ \hline
    \multirow{7}{*}{Reconstruction}&\multirow{3}{*}{Train}&VAE&\textbf{0.03264}&\textbf{0.661}&\textbf{0.38827}&39.61&12.09&\textbf{27.52} \\
    &&VAE+Trx&0.05587&0.53355&0.48241&46.057&10.53&35.52 \\
    &&VAE+AR&0.08704&0.59887&0.51608&48.90747&11.68637&37.22 \\ \cline{2-9}
    &\multirow{4}{*}{Test}&VAE&\textbf{0.03775}&\textbf{0.56436}&\textbf{1.14429}&97.01&14.52&\textbf{82.49} \\
    &&VAE+Trx&0.0791&0.40842&1.38352&142.63245&11.00&131.63\\
    &&VAE+AR&0.08432&0.55446&1.15334&97.82669&14.14&83.68 \\
    &&VAE+Flow&0.03858&0.54703&1.15739&97.73643&14.13&83.59\\
    \hline
    
    \hline
    \multirow{7}{*}{Generation}&\multirow{3}{*}{Train}&VAE&\textbf{0.03899}&\textbf{0.4541}&0.7485&&&  \\
    &&VAE+Trx&0.0616&0.441&\textbf{0.68}&&& \\
    &&VAE+AR&0.10037&0.411&0.80&&& \\ \cline{2-6}
    &\multirow{4}{*}{Test}&VAE&\textbf{0.08838}&\textbf{0.36634}&1.71&&& \\
    &&VAE+Trx&0.10734&\textbf{0.36634}&\textbf{1.62}&&& \\
    &&VAE+AR&0.15394&0.34035&1.76&&& \\
    &&VAE+Flow&0.14517&0.25&2.20&&& \\ \hline
    \end{tabular}
    \end{table}

\section{Analysis}
\begin{itemize}
    \item Baseline VAE model exhibits competitive performance versus the augmented models. 
    \item Baseline VAE performs better on most metrics on training set versus the test set. This could be due to overfitting to the training set. Therefore, we need to implement better regularization techniques. It is even more evident by looking at the generative metrics on the test set, which shows more degradation than reconstruction metrics on the same test set. 
    \item Transformer Encoder in general under-performs the rest of the models. One reason is that we currently apply self-attention across all of the input points. This might not be the best strategy because a point is usually only strong correlated to its neighbours instead of all points. 
    \item Autoregressive decoding improves the generative performance w.r.t to \textbf{JSD}. We currently generate \textbf{256} points in each progressive step along a specific order. Better tuning this hyper-parameter of number of points per group and specific decoding order, it could better exhibit the merits of this design. 
    \item Latent Space Flow model has competitive \textit{reconstruction} performance, however its \textit{generative} metrics are low. We believe the optimization should be improved for this model, as its objective is different from the rest of models.
    \item  Comparing the metrics result among different dataset, it can be observed that chair > airplane > car. One reason can be that car category has more unique 3D shapes. Therefore, if mode collapses into a single one during training, it will definitely perform worse compared to the rest of the dataset during test sample.
    \item There is not a model that perform consistently well across all. One reason can be that each object has its own unique geometric structure, which is better utilized by one of the models.
\end{itemize}

\section{Conclusion}
In this project, we proposed using various ideas to improve the generation and reconstruction performance of VAE models for 3D point clouds. Encoder augmentation by using transformer encoder, latent representation improvement by normalizing flow, and autoregressive decoding were experimented with. We showed that using such techniques could improve the generation and reconstruction performance of the vanilla VAE in some of the metrics.
We are planning to continue this work by following the below ideas:
\begin{itemize}
    \item For encoder, context aware transformer should be used as encoder where each point will look only into its k-nearest neighbors to learn its feature.
    \item For flow, different number of flow layers should be taken into consideration as well as other flow models beside IAF.
    \item For decoder, we should experiment with different number of points per decoding step for the NADE architecture, and also model the decoder using MADE, where each output node of MADE will represents a single point. 
    \item Combining all encoder, latent space, and decoder augmentations.
\end{itemize}

We would like to thank Yang Song, who advised us in this project. We certainly used his insights, and hope to continue to do so. 
Our code is available at https://github.com/lingjiekong/CS236Project.

\bibliography{bibliography}
\bibliographystyle{unsrt}

\end{document}